\documentclass{esannV2}
\usepackage[pdftex]{graphicx}
\usepackage[latin1]{inputenc}
\usepackage{graphicx}
\usepackage{subcaption}
\usepackage[usestackEOL]{stackengine}
\usepackage{amssymb,amsmath,array,natbib}
\usepackage{subcaption}
\usepackage{url}
\usepackage[T1]{fontenc}

\begin{document}

\title{The Stability Gap in \\ Continual Reinforcement Learning}

%***********************************************************************
% AUTHORS INFORMATION AREA
%***********************************************************************

\author{
Sebastian Pusch, Matthia Sabatelli and Gido M. van de Ven
\vspace{.3cm}\\
Department of Artificial Intelligence, University of Groningen\\
Groningen, The Netherlands\\
}

%***********************************************************************
% END OF AUTHORS INFORMATION AREA
%***********************************************************************

\maketitle

\begin{abstract}
Continual reinforcement learning (CRL) aims to enable agents to adapt to changing tasks while preserving previously acquired capabilities. Evaluation often relies on final aggregate metrics that can hide temporary performance degradation during adaptation. This work introduces a simple CRL benchmark that varies observation representations while preserving environment dynamics and rewards, supporting proof-of-concept demonstrations, inexpensive experimentation, and easily reproducible sanity checks. Using this benchmark, we demonstrate the stability gap in CRL: a substantial drop in prior-task performance after the introduction of new tasks, followed by partial or complete recovery. This pattern occurs despite the agent having complete access to past tasks.
\end{abstract}

\section{Introduction}

Sequential training can severely degrade neural network performance on previously learned tasks, a phenomenon known as catastrophic forgetting \citep{MCCLOSKEY1989109}. Continual learning~(CL) aims to adapt to changing tasks while retaining acquired knowledge. Evaluation typically relies on task-boundary measures such as final performance, forgetting, and backward transfer, which can overlook temporary failures during adaptation. In supervised CL, De Lange et al.~\cite{delange2023continualevaluationlifelonglearning} identified the \emph{stability gap}: a sharp drop in previous-task performance after a task transition, followed by partial or complete recovery. This gap can occur even when all past-task data remain available~\citep{hess2024complementaryperspectivescontinuallearning}, a setting referred to as incremental joint-training.

Whether a stability gap also occurs in continual reinforcement learning~(CRL) remains unexplored. Existing CRL evaluations emphasize final performance, or they look at transfer and forgetting measured only at task boundaries or after convergence \citep{NEURIPS2021_ef8446f3}. Temporary performance drops may therefore have remained hidden.
To isolate these dynamics, we introduce a lightweight transformed-state benchmark inspired by Permuted-MNIST \citep{goodfellow2013empirical}. Tasks share environment dynamics, rewards, and action spaces but apply different orthogonal transformations to the state representation. We compare value-based and actor-critic algorithms across discrete and continuous action spaces, using frequent evaluation of past tasks to test whether a stability gap arises in CRL. %In this setting, we observe a stability gap in every algorithm tested, across both discrete and continuous action spaces.
\section{Benchmark}

\begin{figure}[ht!]
  \centering
  
  \stackinset{l}{0pt}{t}{-15pt}{{\footnotesize\textbf{(a)}}}{%
    \includegraphics[scale=1]{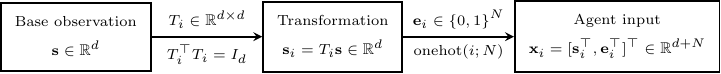}%
  }
  
  \vspace{1.5em}

  \stackinset{l}{0pt}{t}{10pt}{{\footnotesize\textbf{(b)}}}{%
   \includegraphics[width=1\linewidth]{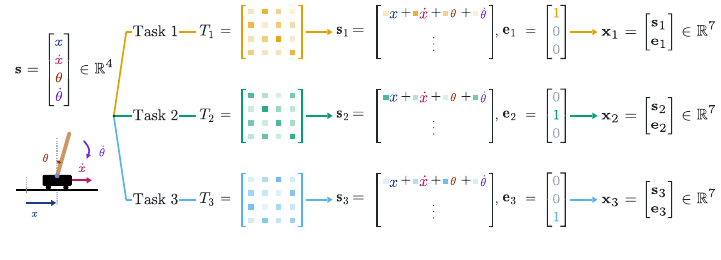}%
  }
  
  \vspace{-0.5em}

  \stackinset{l}{0pt}{t}{-12pt}{{\footnotesize\textbf{(c)}}}{%
    \includegraphics[width=1\linewidth]{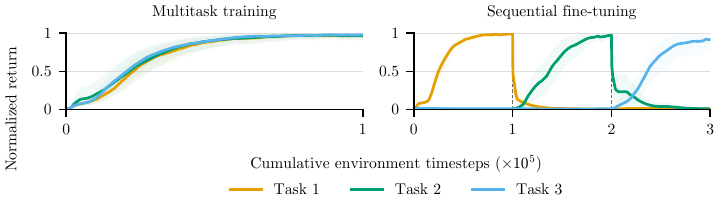}%
  }

  \caption{\textbf{Benchmark construction and CartPole instantiation.} 
  \textbf{(a)}~Task-specific agent inputs are constructed by applying an orthogonal transformation~\(T_i\) to a base state~\(\mathbf{s}\) and concatenating the result with a one-hot task identity. \textbf{(b)}~An instantiation on three CartPole tasks: each four-dimensional state is transformed by a task-specific matrix and paired with a three-dimensional task identifier, yielding a seven-dimensional input. \textbf{(c)}~All three tasks can be learned by multitask training (left), whereas sequential fine-tuning (right) shows catastrophic forgetting. Illustrated here for SAC-D on \texttt{CartPole}.} 
  \label{fig:benchmark}
\end{figure}

We construct a controlled CRL benchmark in a way that can be applied to any environment with vector observations.\footnote{Code and benchmark: \url{https://github.com/sebapusch/stability-gap-drl}} Each task shares the same dynamics, rewards, and action space but uses a different observation representation. For each experimental seed, we independently sample $N$ orthogonal matrices $T_i \in \mathbb{R}^{d \times d}$ and transform the state $\mathbf{s} \in \mathbb{R}^d$ as
\[
\mathbf{x}_i=
\begin{bmatrix}
T_i\mathbf{s}\\
\mathbf{e}_i
\end{bmatrix},
\]
where $\mathbf{e}_i$ is a one-hot task identifier. We obtain $T_i$ by QR decomposition of Gaussian matrices, correcting column signs using the diagonal of $R$. Orthogonality preserves angles, distances, and information, retaining the geometry needed to learn the underlying dynamics. The task identifier can be omitted for a harder variant of the benchmark.
% The task identifier disambiguates observations and removes the need for task inference. Omitting it provides a more challenging variant, which remained learnable for some algorithms in our preliminary experiments. All experiments reported here use task identifiers.

Figure~\ref{fig:benchmark} presents the general benchmark construction (a) and its instantiation on \texttt{CartPole} (b), which we use alongside \texttt{InvertedPendulum} for discrete and continuous control, respectively. Both environments have four-dimensional states ($d=4$), and we construct three tasks ($N=3$) for each. We verified that with our experimental settings, all algorithms we consider can learn all three tasks under multitask training, yet exhibit catastrophic forgetting under sequential fine-tuning. Panel~(c) illustrates this for SAC-D.
\section{Experimental setup}

\paragraph{Algorithms}
We evaluate DQN \cite{mnih2013playingatarideepreinforcement} and discrete SAC (SAC-D) \cite{DBLP:journals/corr/abs-1910-07207} on \texttt{CartPole}, and DDPG \cite{lillicrap2019continuouscontroldeepreinforcement} and SAC \cite{haarnoja2018softactorcriticoffpolicymaximum} on \texttt{InvertedPendulum}. All algorithms are trained and evaluated under an incremental joint-training paradigm: tasks are introduced sequentially, and the agent continues to train on all tasks encountered thus far. Unlike with methods that rely on limited replay or other approximations of prior-task experience, the agent thus retains full access to all past tasks, making any stability gap especially notable.
For all algorithms, we maintain a separate replay buffer for each task and sample equal proportions of transitions from these buffers for each update. At each task transition, we reset the replay buffers and optimizer states while retaining the learned network parameters. Then, we collect 1,000 transitions from each encountered task before resuming network updates. We evaluate performance on all tasks every 15 environment steps during the first 30,000 steps of each task and every 250 steps thereafter. Performance on a task at a given timestep is measured as the mean return over 15 episodes. 

For all experiments, we use the Adam optimizer, networks with two fully-connected hidden layers of 256 units each, a batch size of 128, and a discount factor $\gamma=0.99$. We tune the learning rate separately for each algorithm, selecting the configuration with the highest mean return across all tasks at the end of training, averaged over 10 seeds. Tuning consists of an initial logarithmic sweep followed by up to three additional values near the best-performing region. The selected learning rates are $10^{-4}$ for DQN, $3\times10^{-4}$ for DDPG and SAC, and $10^{-3}$ for SAC-D. We then run the experiments with the selected learning rates on 50 new seeds. Regarding the exploration hyperparameters, for DQN we follow \cite{mnih2013playingatarideepreinforcement} and employ an \(\epsilon\)-greedy policy with an annealing schedule. In our experiments, \(\epsilon\) decreases from 1.0 to 0.05 over the first 10\% of timesteps, after which it remains constant. When it comes to SAC and SAC-D, we use automatic temperature tuning as proposed in  \cite{haarnoja2019softactorcriticalgorithmsapplications, DBLP:journals/corr/abs-1910-07207}, with target entropies of $-d_a$ and $0.98\log|\mathcal{A}|$, respectively. Here, $d_a$ denotes the dimensionality of the continuous action space, and $|\mathcal{A}|$ denotes the number of discrete actions.

\paragraph{Linear Interpolation} For each of the 50 experimental seeds, we evaluate a $200 \times 200$ parameter grid built from the policies \(\theta_1, \theta_2, \theta_3\) learned after Tasks 1--3 in that run. The extrapolated fourth corner is \(\theta_4=\theta_1+\theta_3-\theta_2\), and each grid policy is formed by bilinear interpolation:
\[
\theta(u,v)=v\big[(1-u)\theta_1+u\theta_2\big]+(1-v)\big[(1-u)\theta_4+u\theta_3\big],
\qquad u,v\in[-0.5,1.5].
\]
At each grid point, we evaluate the performance on Task 1 by averaging the return over 15 episodes per seed.

\section{Results}

\begin{figure*}[!t]
    \centering
    \includegraphics[width=1\linewidth]{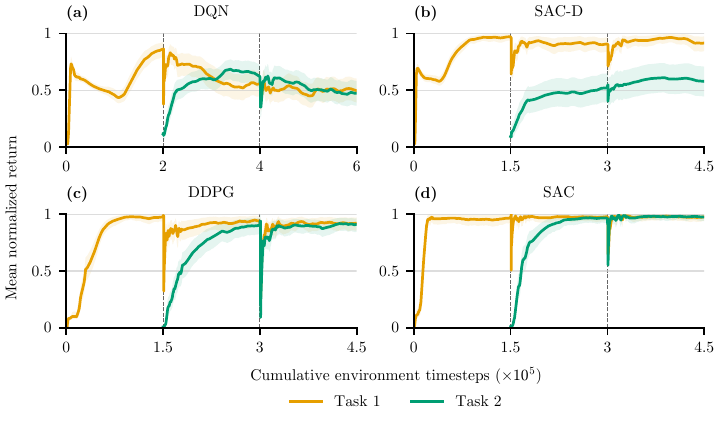}

   \caption{\textbf{A stability gap consistently shows up in CRL.}
    Performance on Task~1 and 2 during incremental joint-training on three tasks for \textbf{(a)} DQN, \textbf{(b)}~SAC-D, \textbf{(c)} DDPG, and \textbf{(d)} SAC. Dashed vertical lines mark task transitions. Curves show mean normalized returns across seeds, with shaded 95\% bootstrapped confidence intervals, revealing transient drops in past-task performance in all settings.
   }

    \label{fig:learning_curves}
\end{figure*}

\begin{figure}[!t]
    \centering
    \includegraphics[width=1\linewidth]{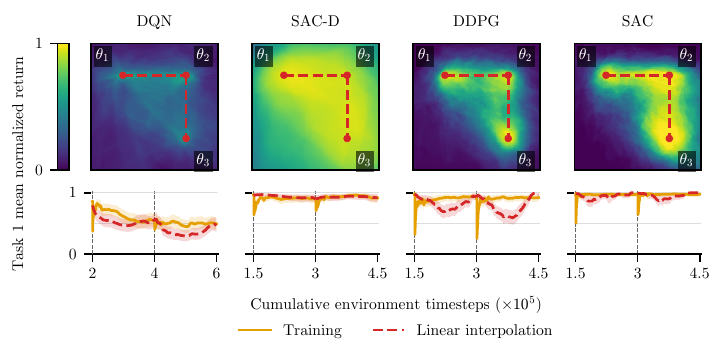}
    \caption{\textbf{Training trajectories tend to boost larger performance drops than corresponding linear paths.} Task 1 return along a three-checkpoint parameter-space path for DQN, SAC-D, DDPG, and SAC. Top: heatmaps show normalized mean Task 1 return; the red dashed path connects \(\theta_1\), \(\theta_2\), and \(\theta_3\). Bottom: Task 1 return during training on Task 2 and 3 (solid orange) and along linear interpolations between successive checkpoints (dashed red), with shaded 95\% bootstrapped confidence intervals. Interpolation coefficients are mapped linearly onto the corresponding training intervals for visualization. Interpolation produces pronounced return drops for DQN and DDPG, while SAC-D and SAC remain near their checkpoint returns.}
    \label{fig:heatmaps}
\end{figure}

All four algorithms exhibit a sharp, substantial drop in previous-task performance after at least one task transition, followed by partial or complete recovery~(Figure~\ref{fig:learning_curves}).
Control experiments confirm that this pattern persists when optimizer states and replay buffers are retained across task transitions, when exploration parameters are reset at task transitions, and when returns are aggregated across seeds using the interquartile rather than the arithmetic mean.

Figure~\ref{fig:heatmaps} compares Task~1 return along linear interpolations between successive checkpoints with Task 1 return during training. For SAC-D and SAC, Task~1 returns along the interpolated paths are, on average across seeds, close to those at the checkpoints, even though training shows pronounced transient drops. Thus, stable high-return linear paths between checkpoints can coexist with a stability gap along the training trajectory. For all algorithms except DQN, the dips in mean return during training are larger than the corresponding dips along the interpolated paths.

\section{Conclusion}

We demonstrate the existence of the stability gap in CRL: previous-task performance can drop sharply following a task transition, even under incremental joint-training.
Consistent with prior work on unstable optimization trajectories in supervised CL \cite{yoo2024layerwiseproximalreplayproximal, hess2024complementaryperspectivescontinuallearning}, our finding shows that even an objective that jointly incorporates all tasks encountered thus far is not, by itself, sufficient to maintain stable performance throughout training. The optimization trajectory also matters. This motivates future work on CRL methods that change the way optimization is done, particularly during the updates following task transitions, when the stability gap may emerge.

Additionally, we propose a simple benchmark for CRL in which tasks vary only in their observation representations, while rewards and environment dynamics remain fixed. Although this design does not capture the full range of task changes encountered in CRL, its simplicity serves three purposes: (i)~it supports proof-of-principle demonstrations, as in this study; (ii)~it provides a standardized, inexpensive playground for initial iterations on new CRL methods; and (iii)~it offers an easily reproducible sanity check for new algorithms.

\begin{footnotesize}
\bibliographystyle{unsrt}
\bibliography{main}
\end{footnotesize}

\end{document}